\documentclass{article} %

\usepackage{authblk}
\usepackage[numbers]{natbib}
\usepackage{geometry}
\usepackage{amsmath,amsfonts,bm}

\def\eqref#1{equation~\ref{#1}}

\def\1{\bm{1}}

\DeclareMathAlphabet{\mathsfit}{\encodingdefault}{\sfdefault}{m}{sl}
\SetMathAlphabet{\mathsfit}{bold}{\encodingdefault}{\sfdefault}{bx}{n}

\usepackage{hyperref}
\usepackage{url}
\usepackage{graphicx}

\title{A framework for fully autonomous design of materials via multiobjective optimization and active learning: challenges and next steps}
\author[1]{Tyler H.\ Chang\thanks{Correspondence to \href{mailto:tchang@anl.gov}{tchang@anl.gov}}}

\author[2]{Jakob R.\ Elias}
\author[1,3]{Stefan M.\ Wild\thanks{Work was primarily performed at Argonne, prior to changing institutions}}
\author[2,4]{Santanu Chaudhuri}
\author[2]{Joseph A.\ Libera}
\affil[1]{Mathematics and Computer Science Division, Argonne National Laboratory \texttt{tchang@anl.gov}}
\affil[2]{Applied Materials Division, Argonne National Laboratory \texttt{\{jelias,schaudhuri,jlibera\}@anl.gov}}
\affil[3]{Applied Mathematics and Computational Research Division, Lawrence Berkeley National Laboratory \texttt{wild@lbl.gov}}
\affil[4]{Department of Civil, Materials, and Environmental Engineering, University of Illinois Chicago}
\date{April 2023}

\begin{document}

\maketitle

\begin{abstract}
    In order to deploy machine learning in a real-world self-driving laboratory where data acquisition is costly and there are multiple competing design criteria, systems need to be able to intelligently sample while balancing performance trade-offs and constraints.
    For these reasons, we present an active learning process based on multiobjective black-box optimization with continuously updated machine learning models.
    This workflow is built on open-source technologies for real-time data streaming and modular multiobjective optimization software development.
    We demonstrate a proof of concept for this workflow through the autonomous operation of a continuous-flow chemistry laboratory, which identifies ideal manufacturing conditions for the electrolyte 2,2,2-trifluoroethyl methyl carbonate.
    
\end{abstract}

\section{Introduction}
\label{sec:intro}

Data-driven automated laboratories, also called self-driving laboratories, can significantly accelerate molecular synthesis and materials discovery.
A key technical challenge of fully autonomous and artificial intelligence-assisted laboratory design is to effectively collect and utilize data from multiple complex processes in order to inform future experimentation.
One long-standing template for utilizing experimental and simulation data is the multiresponse surface methodology (RSM) \citep{myers2016}, whereby initial data sets are gathered through design of experiments and then statistical models are built for each quantity of interest, analyzed, and hypothesis tested iteratively.

In modern scientific and engineering settings,  several common paradigms could be considered as specific implementations of this discovery framework, the most common of which are active learning and model-based optimization techniques such as Bayesian optimization.
To account for multiple competing criteria, we utilize an active learning framework based in multiobjective optimization, which utilizes surrogates (such as Gaussian processes), optimization solvers, and multicriteria data acquisition in a closed feedback loop.
Several definitions of active learning exist, including both adaptive sampling to model complex processes with uniform accuracy \citep{sapsis2022} and iterative selection of candidate designs to improve a global machine learning model that drives optimization convergence \citep{yuan2023}.
In this paper we use the latter definition, since global model accuracy is not a reasonable goal when using an extremely limited budget for experiments, as in our case.

We present our approach to integrate streaming experimental data from a real-world, fully automated continuous-flow chemistry setup into a customizable machine learning (ML) and multiobjective optimization framework for steering the design of battery electrolytes.
In Section~\ref{sec:bg} we briefly summarize the problem background, focusing on relevant techniques and challenges.
Section~\ref{sec:framework} introduces our framework for addressing these challenges, which is based on integration between the multiobjective optimization library ParMOO \citep{chang2022c} and the data streaming platform for Manufacturing Data and Machine Learning (MDML) \citep{elias2020}.
In Section~\ref{sec:experiment} we present our results from the synthesis of the battery electrolyte 2,2,2-trifluoroetheyl methyl carbonate (TFMC) with an integrated automated feed, continuous-flow reactor (CFR), and nuclear magnetic resonance (NMR) setup.
This experiment does not constitute novel material discovery, but in Section~\ref{sec:futurework} we discuss how our framework will allow for extension to the discovery domain.

\section{Background and Key Challenges}
\label{sec:bg}

There is no shortage of options when it comes to multiobjective optimization solvers and libraries.
Notable techniques include genetic algorithms \citep{blank2020},
search-based methods \citep{ledigabel2011}, and
Bayesian optimization \citep{balandat2020}.
We note that we are not the first paper to take a multiobjective approach to chemical synthesis \citep{hase2018,shields2021}.
However, setting up these solvers to integrate with scientific computing and laboratory environments is a nontrivial task, where each optimization software must be integrated as an optimization service \citep{raghunath2017} or function \citep{chang2020b} in the broader library.
This approach is appropriate for integrating with computer simulation environments; but in order to successfully integrate with heterogeneous data sources, a more flexible approach is needed \citep{barbagonzalez2018}.

Another challenge in applying optimization solvers for autonomous material discovery is exploiting structures to address problem complexity.
Optimization solvers tend to perform well over continuous input spaces; but when some of the design variables are complex (such as chemical networks), the problem becomes combinatorial for classical solvers unless they are carefully tailored to the problem at hand.
In the context of material discovery, the molecular structure can be embedded by using either its molecular descriptors \citep{shields2021} or problem-specific latent-space representations \citep{hoffman2022}.
Similarly, after performing experiments, the raw outputs of these chemical processes may consist of large volumes of time series data or spectral measurements, and the majority of this information is lost when the data is postprocessed to derive meaningful objectives.
However, modern simulation optimization solvers can take advantage of these raw simulation outputs as a method for exploiting the physical structure in the problem \citep{wild2017}.

Thus, our challenge is to exploit problem-specific structures and embeddings in a way that generalizes to a variety of applications while modeling data from multiple chemical processes.
This requires a multiobjective optimization framework that is flexible enough to utilize customized problem definitions, while still leveraging state-of-the-art modeling techniques, and that can be coupled with a flexible data-streaming service.

\section{A Framework for Autonomous Experimentation}
\label{sec:framework}

To address our challenges, we use the multiobjective optimization library ParMOO \citep{chang2022c} to manage multiple surrogate models, exploit domain knowledge, and utilize problem embeddings through its customizable modular framework.
The data-streaming service MDML \citep{elias2020} provides an event-driven architecture using an Apache Kafka instance with publishers and subscribers to pass data between laboratory workstations, servers, supercomputers, and more.
Once an MDML instance has been created---typically on a server separate from experiments and machine learning codes---any number of experiments or clients can begin streaming data.
Additional Kafka instances can be added to provide horizontal scaling if needed.
By utilizing MDML as ParMOO's simulation distribution backend, we are able to stream scientific data from multiple sources, process that data, and issue requests for new experiments in a closed loop.

Additionally, both of these tools are open-source, lightweight Python libraries, which can be installed on a typical laboratory workstation.
To integrate ParMOO with MDML, we have extended ParMOO's {\tt MOOP} class, overwriting the {\tt solve} method to produce Kafka requests for experimental data via MDML.
This means that the MDML platform handles data streaming and recording through its usual methods and is responsible for issuing experiments or simulations, logging data, and storing results.
In this work we used two servers running Kafka brokers as our MDML instance, providing FuncX \citep{chard2020} endpoints for our analysis tasks.
We then used a lab workstation running LabVIEW (which directly connects to the CFR and NMR) and a compute node running a ParMOO solver instance as analysis client processes.
Figure~\ref{fig:mdml-architecture} in Appendix~\ref{sec:bg:mdml} further illustrates this setup.

For more information on how ParMOO facilitates exploiting problem structure, the specific structure and components of its modeling and candidate selection, and the techniques used in this paper, see Appendix~\ref{sec:bg:parmoo}.
For more information on how MDML handles data streaming from heterogeneous sources, see Appendix~\ref{sec:bg:mdml}.

\section{Chemical Synthesis with ParMOO and MDML}
\label{sec:experiment}

In this section we explore electrolyte production in an autonomously operated CFR using the framework from Section~\ref{sec:framework}.
Compared with traditional batch processing,
CFRs can provide a means to rapidly test reagent combinations, experiment with high reaction temperatures, and measure yield and specificity when equipped with online characterization.
However, manual operation can be very expensive, time-consuming, and inefficient.
Thus, automating all aspects of the experimentation process is attractive in order to maximize equipment utilization and minimize labor.
Furthermore, when coupled with ML algorithms, the total number of experiments required to reach optimal parameters is also minimized, in comparison with a traditional exhaustive search and experimental design techniques.

\begin{figure}[h]
    \centering
    \includegraphics[width=0.96\textwidth]{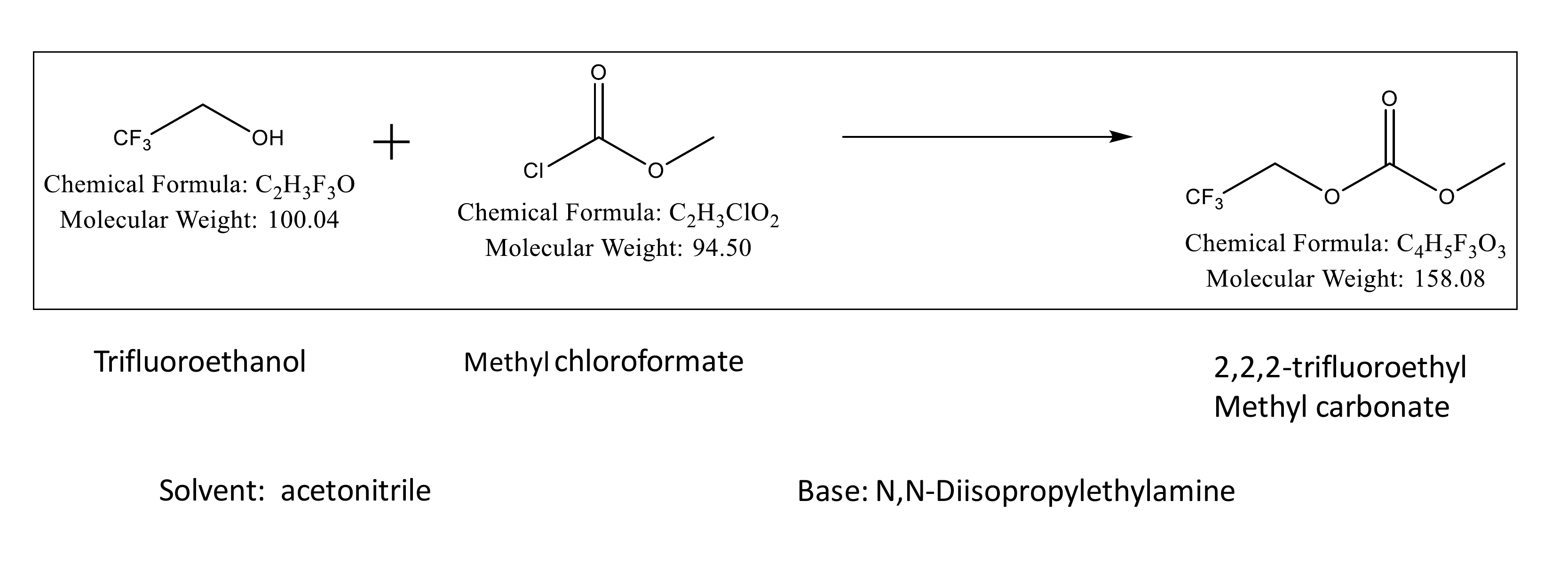}
    \vskip -18pt
    \caption{Diagram of the basic reaction optimized for the production of the electrolyte 2,2,2-trifluoroethyl methyl carbonate (TFMC).}
    \label{fig:reaction}
\end{figure}

For this example we are interested in identifying ideal manufacturing conditions for the battery electrolyte TFMC in a CFR,
based on the reaction illustrated in Figure~\ref{fig:reaction}.
Note that in this example we are using a predetermined pair of reagents, solvent, and base and a limited range of values for the reaction conditions.
Specifically, we must select an optimal flow rate, reaction temperature, and equivalence ratio for the two reagents.
To reduce the reaction time and increase production speed, we would like the reaction to be carried out at high temperatures.
However, high temperatures can also activate a side reaction that produces an unwanted byproduct and reduces the purity of the product.
Therefore, we are seeking chemical mixtures and conditions that produce large amounts of TFMC but small amounts of the byproduct trifluoroethanol (TFE).

To summarize our self-driving laboratory setup,
we created an MDML client process  running a modified ParMOO solver, as described in Section~\ref{sec:framework},
which produced requests for experiments and consumed experimental results.
Specific ParMOO solver settings are given in Appendix~\ref{sec:bg:parmoo}, and specific MDML settings are given in Appendix~\ref{sec:bg:mdml}.
Experiment requests were consumed by a laboratory workstation running LabVIEW, which operates a multiport VICI valve controlling the flow of reagents into a Vaportec CFR.
Outputs from the CFR were shunted into a Magritek NMR, which recorded peak areas for both the product (TFMC) and byproduct (TFE).
The operation of this setup and collection of NMR data were controlled by a workstation running a control program written in LabVIEW and returned to the Kafka brokers via an MDML producer.
Additional details on this lab setup are given in Appendix~\ref{sec:experimental-setup}.

In our case the correspondence of each peak to TFMC or TFE is determined by the ordering of the peaks, which had been identified as a suitable strategy for this experiment via prior experimentation and domain knowledge.
Ultimately, only the peak areas are used in our objective calculations.
However, this reduction sacrifices valuable information about the shape and location of each peak, which could potentially be used to improve surrogate model accuracy \citep{wild2017}.
Additionally, we acknowledge that this might not be a suitable strategy in the general case.

\subsection*{Experiment: Optimal Manufacturing Conditions}
\label{sec:exp1}

Although one can experiment with different options, we have fixed the solvent and base for this experiment to a predetermined pairing that has shown promising results.
We focus on identifying optimal manufacturing conditions, such as flow rates, reaction times, and reaction temperatures.
In order to ensure safe operations of the CFR, upper and lower bound constraints were given for each of these values, as shown in Table~\ref{tab:exp1-bounds}.

\begin{table}[h!]
    \centering
    \begin{tabular}{c|cc}
        Parameter & Lower bound & Upper bound \\
        \hline
         Temperature (degrees C) & 40 & 150 \\
         Reaction time (seconds) & 60 & 300 \\
         Equivalence ratio (no units) & 0.9 & 2 \\
    \end{tabular}
    \caption{Design variables and bound constraints for initial experiment.}
    \label{tab:exp1-bounds}
\end{table}

One of the key challenges for this problem is to achieve convergence on an extremely limited budget.
Since we are running real experiments on a CFR, each experiment has significant costs in terms of financial cost, resources, and time.
As examples,
the raw materials (reagents, solvents, and bases) used in these experiments must have high purity, which comes at a significant financial cost;
the machinery (CFR and NMR) have operating costs;
and the experiments themselves are time-consuming, requiring up to 10 minutes each.
Therefore, we cannot afford to perform hundreds of experiments in this setting, as would be typical in a Bayesian optimization feedback loop.

To directly control the balance between exploration and exploitation and encourage significant exploitation on a limited budget (while uncertainties are still high in some regions of the space), we favor an active learning implementation that is more similar to a traditional statistical RSM, where an initial design of experiments of predetermined size is allocated to drive exploration and initial model accuracy and then all future budget is allocated to exploiting our model, similarly as in \cite{chang2022a}.
We note that, in this strategy, global convergence is determined by the density of the initial design of experiments and is difficult to guarantee for an unknown function.
However, this approach is also more flexible for extension to domain-specific surrogate modeling techniques, which will aid in the transition to chemical discovery \citep{yuan2023}.

\begin{figure}[h!]
    \centering
    \includegraphics[width=0.9\textwidth]{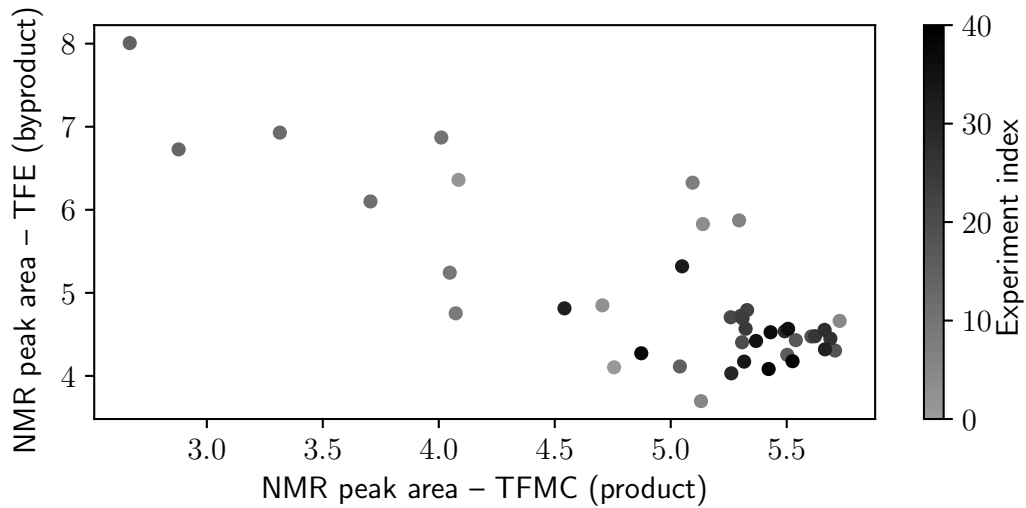}
    \vskip -6pt
    \caption{Product (TFMC) vs.\ byproduct (TFE) areas recorded by NMR for 41 continuous-flow chemistry experiments steered by ParMOO using MDML.
    Point colors are coded by experiment index, with larger indices indicating later runs.}
    \label{fig:exp1-results}
\end{figure}

ParMOO was instructed to maximize the product and minimize the byproduct subject to the constraints in Table~\ref{tab:exp1-bounds}.
After a 15-point initial exploratory design of experiments, ParMOO was run for 9 additional model-exploiting iterations with a batch size of three experiments per iteration (i.e., 27 additional iterative experiments driven by the optimization framework).
Because of lack of continued improvement over several batches of experiments, not all experiments allocated in the budget were carried to completion.
In total, 41 experiments were completed; the corresponding data is shown in Figure~\ref{fig:exp1-results}.
Since we are attempting to maximize the product (TFMC) and minimize the byproduct (TFE), the ideal solution would be located in the bottom-right corner.
Note that the clustering of late (high index) runs in the bottom-right corner indicates convergence to optimal manufacturing conditions with an extremely limited budget.

\section{Path Forward}
\label{sec:futurework}

In Section~\ref{sec:experiment} we were successful in optimizing the manufacturing conditions for a predetermined material, but additional work is needed in order to broaden our scope to the original goal of material {\sl discovery}.
We fully anticipate that the code will continue to find more applications in both synthesis and discovery of materials.
The fundamental method is fully extendable to multireactor, multistep reactions, with more complicated hardware configurations.

As an example, we can use a similar CFR setup to synthesize more complex materials such as metal-organic frameworks (MOFs), with the nucleation and connection of organic linkers to metal nodes being fully visible to the in-line NMR.
However, the detection limits and reaction conditions for this problem generally take months to fully explore and optimize.
The size of MOF libraries and linkers is increasing, but the convergence rates of high-throughput synthesis optimization tools are lacking.
By utilizing a similar autonomous framework, we can extend these methods to reactors for making variations of linkers and MOFs and exploring their synthesis.

This is a much larger problem than the one that has been described in the paper.
To extend to this problem, we will need to utilize domain-specific latent space embeddings, structure-exploiting optimizers that consider the full spectrum of NMR data during surrogate modeling, and heterogeneous data sources including  simulation (for early exploration) followed by experimental data (for late-stage testing).
In order to keep costs reasonable and extend our limited budget, it could also be interesting to consider the cost of each experiment in the acquisition function.
However, we must be careful when doing this in order to still allow for expensive but insightful experiments to be evaluated.

\subsection*{Acknowledgments}
This material was based upon work supported by the U.S.\ Department of
Energy, Office of Science, Office of Advanced Scientific Computing
Research, Applied Mathematics and SciDAC programs under Contract Nos.\
DE-AC02-05CH11231 and DE-AC02-06CH11357 
and by the Argonne LDRD program. 

\bibliographystyle{iclr2023_conference}
\bibliography{moo-refs} %

\appendix
\section{Software Details}

\subsection{ParMOO}
\label{sec:bg:parmoo}

\begin{figure}[h!]
\label{fig:des-sim-obj}
\begin{center}
\includegraphics[width=0.75\textwidth]{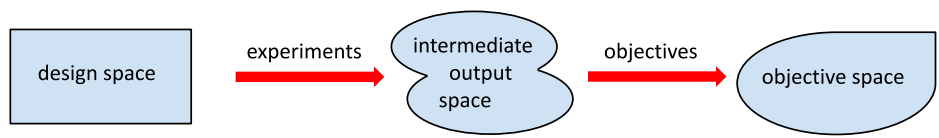}
\end{center}
\caption{ParMOO models complex processes separately from objectives, in order to exploit physical structure in how the objectives are defined.}
\end{figure}

ParMOO is a framework for customizing and deploying multiobjective simulation optimization solvers \citep{chang2022c}.
The key distinction in ParMOO is the difference between a {\sl simulation} and an {\sl objective}, as depicted in Figure \ref{fig:des-sim-obj}.
This abstraction and the utilization of custom latent-space embedding tools are the primary mechanisms by which ParMOO exploits problem structure.

To create a ParMOO instance, users must first specify all design variables, data sources (e.g., experiments and simulations), problem constraints, and problem objectives.
For complex design variable types, ParMOO is capable of automatically generating latent space embeddings or utilizing customized embedding tools.
Next, users provide
\begin{itemize}
\item any pre-existing data or a technique for generating an initial design of experiments;
\item surrogate models for learning each of the simulation outputs;
\item optimization solvers for the surrogate problem; and
\item one or more data acquisition functions for determining how ParMOO will blend objectives.
\end{itemize}

\begin{figure}
    \centering
    \includegraphics[width=\textwidth]{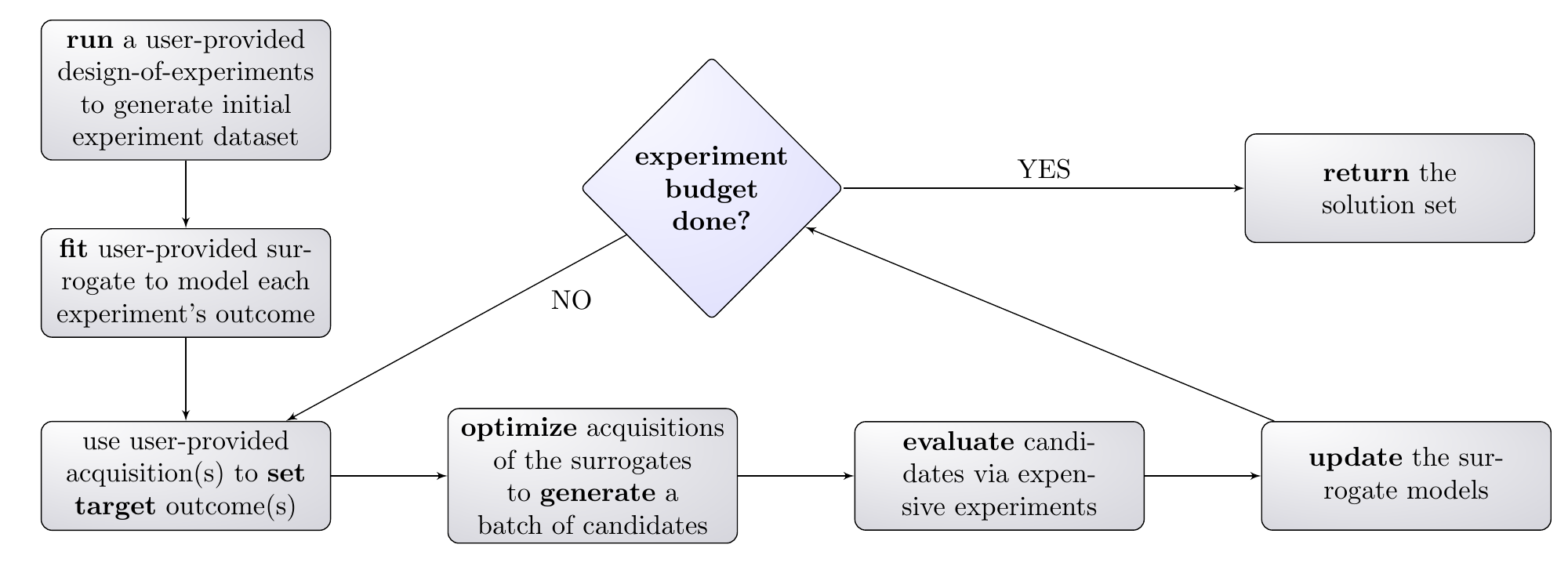}
    \caption{Program control flowchart, depicting how ParMOO combines user-provided techniques to optimize complex processes via active learning.}
    \label{fig:parmoo-flowchart}
\end{figure}

When run, ParMOO will evaluate the initial experimental design, then iteratively generate batches of new experiments to be distributed for evaluation across all data sources, by following the standard response surface methodology.
This feedback loop is illustrated  in Figure~\ref{fig:parmoo-flowchart}.
For further information on this process, see ParMOO's online documentation \citep{chang2022b}.

For the demonstration described in Section~\ref{sec:experiment}, ParMOO was configured to use a 15-point Latin hypercube design of experiments and a Gaussian RBF (the mean function for a Gaussian process) surrogate, and the surrogate problem was solved via generalized pattern search.
Three acquisition functions were provided (resulting in candidate batches of size three), two of which used the epsilon-constraint method to scalarize the problem and the third of which used a fixed 50-50 weighting of the two objectives (i.e., maximize the product and minimize the byproduct).

\subsection{MDML}
\label{sec:bg:mdml}

The MDML platform provides cyberinfrastructure to standardize the research and operational environment to act on scientific data streams and integrate ML in the loop to steer or optimize experiments.
MDML has been designed to meet the needs of in situ measurements for accelerating scalable materials manufacturing while providing capabilities that can be easily adapted to any scientific domain.
MDML enables users to construct rich, data-oriented analysis pipelines that span disparate computational environments.

Underlying MDML is the widely used, distributed event streaming platform named Apache Kafka.
This allows scientists using  MDML to follow an event-driven architecture. 
Clients connected to the MDML's Kafka service are able to publish and subscribe to one or many data channels that they have created.
Clients here include researchers themselves, software programs controlling/monitoring experiments, and ML models.
Given this architecture, any relevant experiment data can be streamed anywhere and fed into client processes running ML models (such as ParMOO) in a unified format.
Additionally, when the experimental data collection process can be fully automated (as with our integrated feed, CFR, and NMR setup), this enables a user-free feedback loop.

\begin{figure}[h!]
    \centering
    \includegraphics[width=\textwidth]{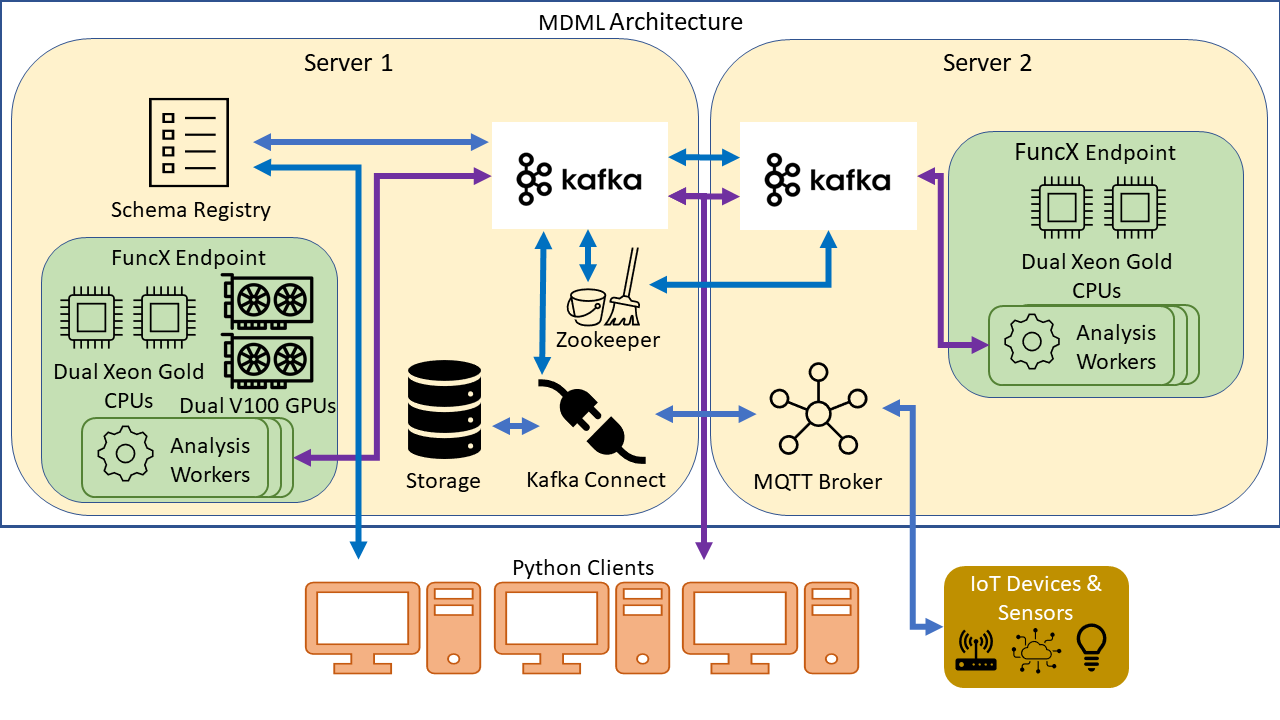}
    \caption{Diagram of the MDML architecture spread across two servers. Purple arrows represent data streams from producers and consumers.}
    \label{fig:mdml-architecture}
\end{figure}

The architecture of our MDML instance utilizes two servers to run two Kafka brokers and their various components, databases for long-term storage of streaming data, and FuncX's \citep{chard2020} 
serverless compute endpoints for spawning analysis jobs that act on streamed data.
A diagram of this architecture is shown in Figure~\ref{fig:mdml-architecture}.

In the diagram, the CFR is illustrated as a Python client since CFR data is collected locally via LabVIEW and streamed out via a Python script.
The CFR also receives suggested experiments from ParMOO via a Python client and loads this information into LabVIEW. 
Here ParMOO is represented as an analysis worker.
It receives results published from the CFR client, decides which experiment should be performed next, and publishes its own data, which is received by the CFR, completing the loop. 
The last important component to note here is Kafka, which is responsible for ushering produced data messages to the appropriate data consumers (i.e., CFR results to ParMOO and ParMOO's suggested experiments to CFR). 
Several MDML features shown in this diagram were not used for the automation of the CFR. 
These include the MQTT broker for collecting data from IoT devices and sensors; databases for storage; and Kafka Connect, which automatically converts data from the latter two sources to Kafka topics and vice versa.

\section{Experimental Methods}
\label{sec:experimental-setup}

TO perform autonomous experimentation on the CFR using ParMOO and the MDML in a closed feedback loop, we created
a MDML client process running a modified ParMOO solver, as described in Section \ref{sec:framework}.
Next, we created an additional experiment-running client  on a laboratory workstation that directly controlled the CFR.
Then, the MDML host was run on a remote laboratory mainframe to broker requests and collect data.

As  described in Section \ref{sec:bg:parmoo}, the ParMOO client consists of a ParMOO solver that was configured to use a 15-point Latin hypercube design of experiments, Gaussian RBF surrogate model, generalized pattern search optimizer, and three acquisition functions (two using the epsilon-constraint method for Pareto front exploration and the third using a fixed weighting of the two objectives).
Following the active learning framework, the ParMOO client generated an initial design of experiments and then posted Kafka producer requests for these designs, listened for results data through a Kafka consumer, used those results to update its Gaussian models, optimized those models to target new experimental designs, and then posted new producer requests in an iterative feedback loop.
In order to reduce experiment times on the CFR, experiment requests within a single batch were sorted by the reaction temperature.
This process was beneficial since it takes additional time to stabilize temperatures when adjusting reaction temperatures by large amounts.

For the experiment client, a lab computer running the LabVIEW control program (LVCP) was configured to consume Kafka requests from the ParMOO client.
The LVCP consumer polled for a producer request in the form of a valid experiment parameter test set.
Whenever a valid experiment parameter test set was received from the ParMOO client, an experiment was set in motion by automated deployment of the provided set on the CFR and NMR setup, described in the next paragraph.
After completion of a test, the LVCP posted the result as a Kafka producer and awaited the next data point as a consumer.

\begin{figure}[h]
\centering
\includegraphics[width=0.4\textwidth]{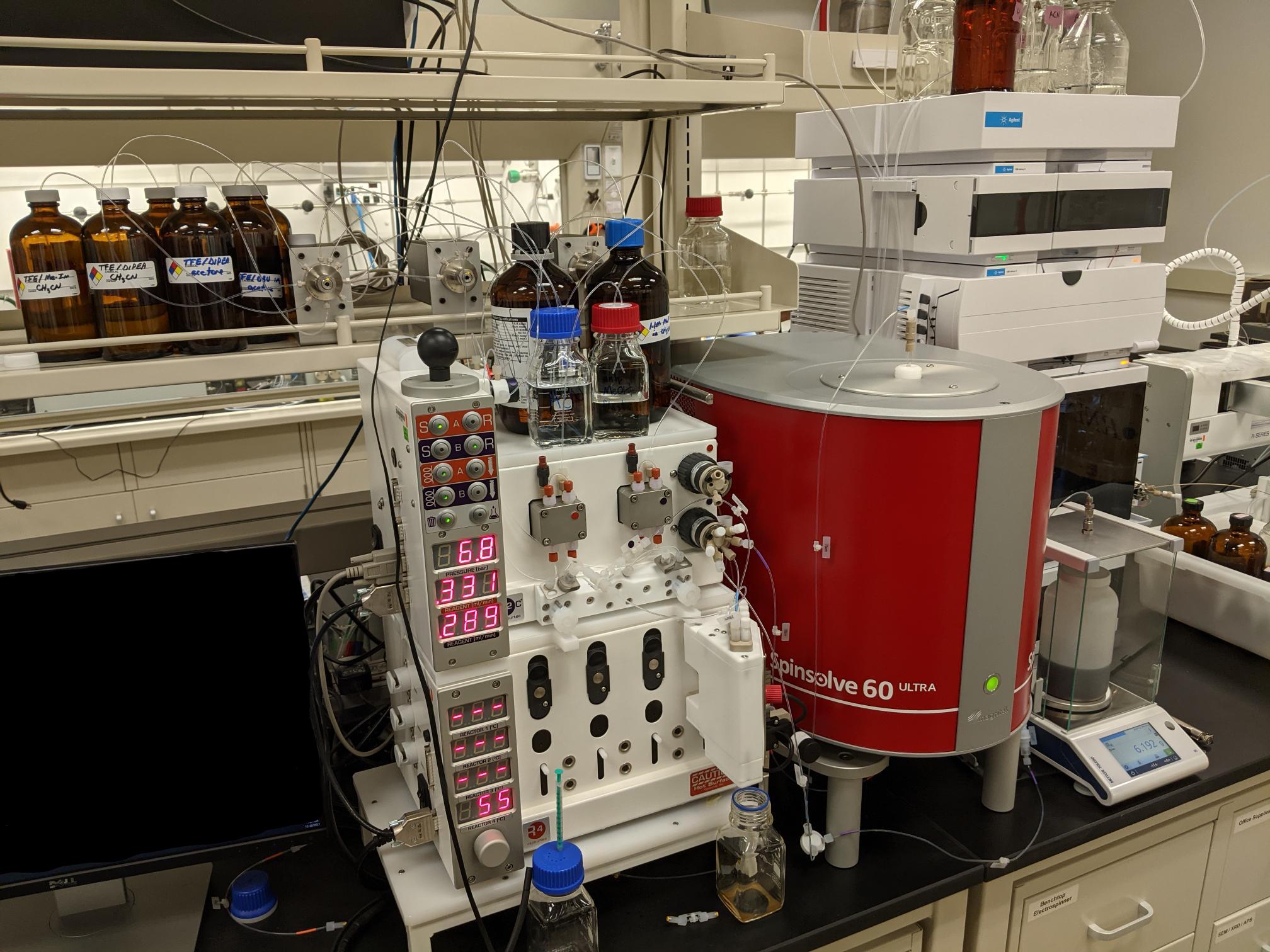}
\caption{Our laboratory setup for performing the physical experiments.
From left to right on the workbench are the lab workstation running the LVCP, the Vaportec CFR unit with automated feed, and the Magritek NMR recording results.}
\label{fig:labsetup}
\end{figure}

In order to perform the physical experiments, a CFR apparatus manufactured by Vaportec was used to carry out continuous-flow reactions.
Inputs to the A and B channels of the Vaportec unit were provided by multiport VICI valves connecting up to 15 possible reagents to the CFR.
The output from the CFR was shunted through a Magritek NMR equipped with a flow-through cell.
The selector valves, Vaportec reactor, and Magritek NMR were fully controlled by using a Labview control program.
A photograph illustrating this setup is given in Figure~\ref{fig:labsetup}.

In order to calibrate the equipment, the reactor was manually set up and NMR spectra tests were run on pure solvents.
Operating parameter ranges were also set.
Whenever a valid experiment parameter set was received, the experiment was initiated from LVCP.
Steady state was determined by monitoring the integrated NMR peak areas provided by the Magritek software, and test completion was determined by the integrated peak area falling below a standard deviation criterion set in advance.
To conclude a single experiment, data acquisition by the NMR required stopping of flow though its flow-through cell, whereas it was not possible to stop the flow in the CFR.
Therefore a bypass loop was employed on the CFR in order to terminate each experiment.

\section{Open Access Statement}

We acknowledge that it will not be possible for researchers to reproduce our results without access to highly specialized laboratory equipment.
However, in order to facilitate open access and reproducibility of results as much as possible, the source code for using ParMOO to create MDML consumer/producer services and a variation of our script used for Section~\ref{sec:exp1} (demonstrating solver settings, but not running the experiments) is publicly available at \url{https://github.com/parmoo/cfr-materials}.
We have also included a copy of the raw MDML-recorded outputs (a JSON file) from our experiment, as well as a Python script for postprocessing these results to produce the plot in Figure~\ref{fig:exp1-results}.

\begin{flushright}
\scriptsize
\framebox{\parbox{\textwidth}{
The submitted manuscript has been created by UChicago Argonne, LLC, Operator of Argonne National Laboratory (“Argonne”).
Argonne, a U.S. Department of Energy Office of Science laboratory, is operated under Contract No. DE-AC02-06CH11357.
The U.S. Government retains for itself, and others acting on its behalf, a paid-up nonexclusive, irrevocable worldwide
license in said article to reproduce, prepare derivative works, distribute copies to the public, and perform publicly
and display publicly, by or on behalf of the Government.  The Department of Energy will provide public access to these
results of federally sponsored research in accordance with the DOE Public Access Plan.
\url{http://energy.gov/downloads/doe-public-access-plan}
}}
\normalsize
\end{flushright}

\end{document}